\documentclass{svproc}
\usepackage{url}
\usepackage{makecell}

\usepackage[utf8]{inputenc}
\usepackage[T1]{fontenc}
\usepackage{lmodern}
\usepackage{microtype}

\usepackage{amsmath,amssymb}

\usepackage{graphicx}
\usepackage{subcaption}
\usepackage{booktabs}
\usepackage{siunitx}
\usepackage{multirow}
\usepackage{array}
\usepackage{threeparttable}
\usepackage{xcolor}

\usepackage[authoryear,round]{natbib}
\usepackage[hidelinks]{hyperref}
\usepackage{cleveref}

\newcommand{\deltacoq}{\Delta_{\text{cog}}}

\newcommand{\accmem}{\alpha_{\text{mem}}}
\newcommand{\accreas}{\alpha_{\text{reas}}}

\begin{document}
\mainmatter              
\title{Dropout Robustness and Cognitive Profiling of Transformer Models via Stochastic Inference}
\titlerunning{Dropout Robustness of Transformers}  
%
\author{Ant\^{o}nio Junior Alves Caiado\inst{1} \and Michael Hahsler\inst{1}}
\authorrunning{A. J. A. Caiado and M. Hahsler} 
%
\tocauthor{Ant\^{o}nio Junior Alves Caiado and Michael Hahsler}
\institute{Lyle School of Engineering, Southern Methodist University, Dallas, TX, USA\\
\email{acaiado@smu.edu, mhahsler@lyle.smu.edu}\\
Corresponding author: acaiado@smu.edu}

\maketitle              

\begin{abstract}
Transformer-based language models are widely deployed for reasoning tasks, yet their behavior under inference-time stochasticity remains underexplored. While dropout is commonly used during training, its effects during inference through Monte Carlo sampling have not been systematically evaluated across diverse architectures, limiting understanding of model reliability in uncertainty-aware applications.

This work conducts a systematic analysis of dropout-induced variability across 19 transformer models using MC Dropout with 100 stochastic forward passes per sample. Dropout robustness is defined as a model's ability to maintain both high accuracy under stochastic inference and stable predictions across dropout samples, measured by standard deviation of per-run accuracies. A cognitive decomposition framework disentangles performance into memory and reasoning components. Experiments span five dropout configurations yielding 95 unique evaluations on 1,000 test samples.

Results reveal substantial architectural variation in dropout robustness. Smaller models demonstrate perfect prediction stability while medium-sized models exhibit notable volatility. Mid-sized models achieve the best overall performance, while larger models excel at memory tasks. Critically, 53 percent of models suffer severe accuracy degradation under baseline MC Dropout, with task-specialized models losing up to 24 percentage points, indicating MC Dropout unsuitability for uncertainty quantification in these architectures. Asymmetric dropout effects emerge: high dropout reduces memory accuracy by 27 percentage points while reasoning accuracy degrades only 1 point, suggesting memory tasks rely on stable representations that dropout disrupts. Systematic cognitive specialization emerges, with 84 percent of models demonstrating memory-biased performance.

This work provides the first comprehensive MC Dropout benchmark for transformers, revealing that dropout robustness is architecture-dependent and uncorrelated with scale. The cognitive profiling framework offers actionable guidance for model selection in uncertainty-aware applications.
\end{abstract}
\section{Introduction}
\label{sec:introduction}
Transformer-based language models have achieved strong performance across natural language processing tasks \cite{vaswani2017attention}. Yet standard evaluations typically report accuracy under \textbf{deterministic inference} (with dropout disabled), providing limited insight into prediction stability under inference-time perturbations. For high-stakes or uncertainty-aware deployment, it is important to understand how model predictions change when stochasticity is introduced at test time, rather than only measuring point estimates of accuracy.

Dropout was introduced as a training-time regularizer \cite{srivastava2014dropout}, and \cite{gal2016dropout} showed that enabling dropout at inference time---Monte Carlo (MC) Dropout---can approximate Bayesian inference by using repeated stochastic forward passes to estimate uncertainty. In transformer architectures, dropout commonly affects both attention and feed-forward sublayers, suggesting that inference-time stochasticity may influence models differently depending on architecture and task demands.

Despite growing interest in transformer reliability, key limitations persist in existing work. Many uncertainty estimation studies focus on narrow model families \cite{desai2020calibration}, and benchmark suites such as GLUE and SuperGLUE aggregate heterogeneous tasks \cite{wang2018glue,wang2019superglue}, obscuring whether robustness differs between factual recall and reasoning. In addition, while scaling laws suggest larger models may be more robust \cite{kaplan2020scaling}, robustness under stochastic inference has not been systematically tested across diverse transformer architectures.

We address these gaps with a large-scale evaluation of MC Dropout robustness across 19 transformer models spanning encoder-only (BERT, RoBERTa, DeBERTa, ELECTRA) and decoder-only (GPT-2, GPT-Neo) families. Each model is evaluated under five dropout configurations that vary attention and feed-forward dropout rates, including deterministic inference and high-dropout settings. To isolate cognitive demands, we introduce a dual-task evaluation that separates \textbf{memory} performance (SQuAD v1.1; \cite{rajpurkar2016squad}) from \textbf{reasoning} performance (HellaSwag; \cite{zellers2019hellaswag}). This design enables us to answer three research questions: (1) How does dropout robustness vary across transformer architectures? (2) Do models exhibit systematic specialization in memory versus reasoning? (3) Does model scale predict stability under stochastic inference?

Our results show that inference-time dropout does not uniformly improve performance, and its impact depends on both architecture and task type. We observe consistent differences between memory and reasoning behavior under high stochasticity, and we find that overall robustness is not explained by scale alone. Together, these findings motivate a more nuanced view of transformer reliability under stochastic inference and provide practical guidance for model selection when uncertainty estimation is required.

\textbf{Contributions.} We (i) provide a broad MC Dropout benchmark across diverse transformer families and dropout configurations, (ii) propose a cognitive decomposition framework separating memory and reasoning evaluation, and (iii) characterize architecture- and task-dependent robustness patterns that are not captured by aggregate benchmark scores.

\section{Related Work}
\label{sec:related}
\subsection{Dropout and Regularization}

Dropout was introduced to reduce overfitting by randomly deactivating neurons during training \citep{srivastava2014dropout}. Variants extend the idea to connections (DropConnect \citep{wan2013regularization}), recurrent states (Zoneout \citep{krueger2017zoneout}), and structured regions (DropBlock \citep{ghiasi2018dropblock}). While these methods target training-time generalization, our focus is inference-time behavior: how models respond when dropout is re-enabled as stochastic perturbation via Monte Carlo Dropout.

\subsection{Monte Carlo Dropout and Uncertainty Estimation}

MC Dropout applies dropout at inference time and uses repeated stochastic forward passes to approximate Bayesian inference and estimate uncertainty \citep{gal2016dropout, kendall2017uncertainties}. It has been applied across vision and NLP tasks, alongside alternatives such as deep ensembles and calibration methods \citep{lakshminarayanan2017simple, guo2017calibration}. Ovadia et al. (2019) evaluated uncertainty methods under dataset shift and found no single approach dominates. However, most prior work evaluates limited model families and does not isolate where dropout is applied (e.g., attention vs.\ feed-forward) as a controlled factor. We address this with 19 models and component-specific dropout variations.

\subsection{Transformer Architecture and Mechanistic Interpretability}

Transformers \citep{vaswani2017attention} underpin widely used models (BERT, GPT, RoBERTa, DeBERTa) \citep{devlin2019bert, radford2019language, liu2019roberta, he2021deberta}. Mechanistic studies suggest component specialization: circuit-level analysis \citep{elhage2021mathematical}, induction heads \citep{olsson2022context}, feed-forward layers as memory-like key--value stores \citep{geva2021transformer}, and attention-head specialization. Motivated by this, we test whether attention and feed-forward sublayers exhibit different robustness under inference-time stochasticity by varying dropout in each component independently.

\subsection{Model Robustness and Reliability}

Reliability beyond accuracy includes adversarial and distribution-shift robustness and confidence calibration. Prior work shows transformer vulnerability to small input perturbations \citep{jin2020bert} and reports robustness gains from pre-training under natural shifts \citep{hendrycks2020pretrain}. Calibration studies find that higher accuracy does not guarantee well-calibrated confidence \citep{desai2020calibration, shelmanov2021label}. Scaling laws suggest larger models should generalize better \citep{kaplan2020scaling, hoffmann2022training}, but whether scale predicts robustness under stochastic inference remains unclear. We test this directly and observe that architecture is a stronger predictor than parameter count.

\subsection{Cognitive Task Decomposition in Language Models}

Task decomposition helps separate factual recall from reasoning. SQuAD emphasizes extracting factual answers from context \citep{rajpurkar2016squad}, while HellaSwag targets commonsense continuation selection \citep{zellers2019hellaswag}; related benchmarks probe multi-step reasoning \citep{talmor2019commonsenseqa}. Probing work suggests different layers encode different linguistic information \citep{tenney2019bert, rogers2020primer}, but these lines of work do not systematically compare memory vs.\ reasoning across architectures under controlled stochastic inference. Our dual-task setup makes this distinction explicit and reveals consistent specialization patterns across model families and dropout settings.

\subsection{Positioning Our Contribution}

Prior work has studied dropout as regularization \citep{srivastava2014dropout}, MC Dropout for uncertainty estimation \citep{gal2016dropout}, mechanistic transformer analyses \citep{elhage2021mathematical}, and robustness/calibration in NLP \citep{hendrycks2020pretrain, desai2020calibration}. However, no study systematically evaluates inference-time dropout robustness across diverse transformer architectures while (i) independently varying dropout in attention vs.\ feed-forward layers and (ii) separating evaluation into factual recall versus commonsense inference using standard benchmarks (SQuAD and HellaSwag) \citep{rajpurkar2016squad, zellers2019hellaswag}.

\section{Methodology}
\label{sec:methodology}
We fine-tune and evaluate 19 transformer models on a balanced two-domain dataset: 500 memory items (factual recall from SQuAD) and 500 reasoning items (commonsense from HellaSwag). After training, each model is evaluated under five dropout configurations using Monte Carlo (MC) Dropout at test time, with 100 stochastic forward passes per test sample. This setup supports three controlled comparisons: (i) robustness across architectures and scales, (ii) dropout location effects (attention vs.\ feed-forward), and (iii) cognitive specialization (memory vs.\ reasoning). The following subsections describe dataset construction (\S\ref{subsec:dataset}), model selection (\S\ref{subsec:models}), dropout configurations (\S\ref{subsec:dropout}), training procedures (\S\ref{subsec:training}), evaluation protocols (\S\ref{subsec:evaluation}), and implementation details (\S\ref{subsec:implementation}).

\subsection{Dataset Construction}
\label{subsec:dataset}

\subsubsection{Data Sources}

We use two established datasets with distinct cognitive demands: SQuAD v1.1 \citep{rajpurkar2016squad} for memory and HellaSwag \citep{zellers2019hellaswag} for reasoning. SQuAD emphasizes factual extraction from Wikipedia passages, while HellaSwag requires selecting plausible event continuations in everyday scenarios. This pairing enables a clean separation between factual recall and commonsense inference. Table~\ref{tab:dataset_sources} summarizes dataset characteristics and sampling.

\begin{table}[t]
\centering
\caption{Dataset Sources and Sampling Strategy}
\label{tab:dataset_sources}
\begin{tabular}{lllll}
\toprule
\textbf{Dataset} &
\textbf{Cognitive Domain} &
\textbf{Source} &
\makecell{\textbf{Sample Size}\\\textbf{(Total/Train/Test)}} &
\makecell{\textbf{Max Input}\\\textbf{Length}} \\
\midrule
\makecell[l]{SQuAD\\v1.1}& \makecell[l]{Memory\\(Factual Recall)} & \makecell[l]{Wikipedia\\Passages} & 500 (400/100) & 200 chars \\
HellaSwag  & \makecell[l]{Reasoning\\(Commonsense)} & \makecell[l]{Event\\Scenarios}     & 500 (400/100) & Full context \\
\bottomrule
\end{tabular}
\end{table}

\subsubsection{Memory Task Construction}

SQuAD is originally extractive QA; we convert it to binary classification. For each question--context pair, we construct one positive example (question + correct answer) and one negative example (question + incorrect answer sampled from a different question's answer set). We ensure negatives differ from the correct answer. For input-length consistency, we filter questions exceeding 200 characters and truncate contexts to 200 characters. This yields 500 balanced memory samples (250 True, 250 False).

\subsubsection{Reasoning Task Construction}

HellaSwag provides a context with four candidate continuations (one correct). We form positives by pairing each context with its gold continuation and negatives by pairing the same context with one randomly selected incorrect continuation. Each context--continuation pair is formatted as a binary classification input. This yields 500 balanced reasoning samples (250 True, 250 False).

\subsubsection{Dataset Split}

We use an 80/20 stratified train--test split, stratified by task type, with random seed 42. This produces 800 training samples (400 memory, 400 reasoning) and 200 test samples (100 memory, 100 reasoning), enabling direct comparison of domain-specific metrics (Section~\ref{subsec:evaluation}).

\subsection{Model Selection}
\label{subsec:models}

We evaluate 19 publicly available transformer checkpoints spanning encoder-only and decoder-only families and a broad range of parameter scales (approximately 4M to 355M). The set includes common baselines (BERT, RoBERTa, DeBERTa), size variants (e.g., tiny to large), domain/task-specialized models (e.g., RoBERTa-SQuAD2, SciBERT), and decoder-only models (GPT-2 variants, GPT-Neo). Table~\ref{tab:models} lists all architectures.

\begin{table}[t]
\centering
\caption{Evaluated Transformer Models}
\label{tab:models}
\begin{tabular}{lll}
\toprule
\textbf{Architecture Family} & \textbf{Type} & \textbf{HuggingFace Checkpoint} \\
\midrule
\multirow{3}{*}{BERT} & Encoder & bert-base-uncased \\
 & Encoder & bert-large-uncased \\
 & Encoder & prajjwal1/bert-tiny \\
\midrule
\multirow{3}{*}{RoBERTa} & Encoder & roberta-base \\
 & Encoder & distilroberta-base \\
 & Encoder & deepset/roberta-base-squad2 \\
\midrule
\multirow{2}{*}{DeBERTa} & Encoder & microsoft/deberta-v3-base \\
 & Encoder & microsoft/deberta-v3-small \\
\midrule
\multirow{6}{*}{Other Encoders} & Encoder & albert-base-v2 \\
 & Encoder & google/electra-base-discriminator \\
 & Encoder & distilbert-base-uncased \\
 & Encoder & SpanBERT/spanbert-base-cased \\
 & Encoder & allenai/scibert\_scivocab\_uncased \\
 & Encoder & sentence-transformers/all-MiniLM-L6-v2 \\
\midrule
\multirow{4}{*}{GPT-2} & Decoder & gpt2 (Small) \\
 & Decoder & openai-community/gpt2-medium \\
 & Decoder & distilgpt2 \\
 & Decoder & sshleifer/tiny-gpt2 \\
\midrule
GPT-Neo & Decoder & EleutherAI/gpt-neo-125m \\
\bottomrule
\end{tabular}
\end{table}

\subsection{Dropout Configurations}
\label{subsec:dropout}

Each model is evaluated under five dropout configurations designed to isolate stochasticity in attention versus feed-forward layers. This is motivated by evidence that these subcomponents play different functional roles \citep{geva2021transformer, olsson2022context}. Table~\ref{tab:dropout_configs} defines the configurations.

\begin{table}[t]
\centering
\caption{Dropout Configurations}
\label{tab:dropout_configs}

\renewcommand{\arraystretch}{1}
\setlength{\tabcolsep}{5pt}

\begin{tabular}{llll}
\toprule
\textbf{Configuration} &
\makecell{\textbf{Attention}\\\textbf{Rate}} &
\makecell{\textbf{FFN}\\\textbf{Rate}} &
\textbf{Description} \\
\midrule
Deterministic & 0.0 & 0.0 & Standard inference mode (dropout disabled) \\
Baseline & 0.1 & 0.1 & Typical training-time dropout rates \\
High Attention & 0.6 & 0.1 & High stochasticity in attention mechanisms \\
High FFN & 0.1 & 0.6 & High stochasticity in feed-forward networks \\
High Both & 0.6 & 0.6 & Elevated stochasticity in all components \\
\bottomrule
\end{tabular}
\end{table}

The baseline rate (0.1) follows common transformer defaults \citep{devlin2019bert}. High dropout (0.6) is used as a stress test that substantially increases stochasticity while preserving non-trivial model behavior. By varying attention and FFN dropout independently, we attribute robustness differences to specific subcomponents rather than uniform dropout effects.

\textbf{Implementation:} We use HuggingFace Transformers \citep{wolf2020transformers}. Dropout rates are applied by modifying model configuration fields prior to loading checkpoints. Because parameter names differ by family (e.g., BERT \texttt{attention\_probs\_dropout\_prob} vs.\ GPT-2 \texttt{attn\_pdrop}), we implement a standardized mapping layer to target attention and FFN dropout consistently across all models.

\subsection{Training Procedure}
\label{subsec:training}

All models are fine-tuned for binary classification using AdamW \citep{loshchilov2019decoupled}. We use learning rate $2 \times 10^{-5}$ with 10\% linear warmup and linear decay, batch size 16 for training and 32 for evaluation, gradient clipping at 1.0, and 5 epochs for all runs. The loss is binary cross-entropy. We keep the same schedule across models to avoid confounding comparisons with differing training budgets.

\textbf{Optimization:} We use mixed precision (FP16) via \texttt{torch.cuda.amp} to reduce memory use and accelerate training, enabling larger models to fit within a 12GB VRAM budget.

\textbf{Hardware:} Experiments run on a single NVIDIA GeForce RTX 3060 GPU (12GB VRAM) with an AMD Ryzen 5 5600X CPU. Across all 95 evaluations, total compute is approximately 28 GPU-hours.

\textbf{Reproducibility:} We fix random seed 42 for data splitting, initialization, and training-time stochasticity. MC Dropout evaluation uses independent dropout masks across forward passes.

\subsection{Evaluation Protocol}
\label{subsec:evaluation}

\subsubsection{Monte Carlo Dropout Procedure}

Following \citet{gal2016dropout}, we enable dropout at inference time by placing dropout modules in training mode while keeping the model otherwise in evaluation mode. For each test sample, we perform 100 stochastic forward passes with independently sampled dropout masks, producing a distribution of predictions used to quantify stability.

\subsubsection{Accuracy Calculation}

Rather than averaging probabilities across passes before scoring, we compute run-level accuracy to directly measure prediction variability. For each of the 100 forward passes, we produce predictions for all 200 test samples and compute accuracy for that pass. We then report the mean and standard deviation across the 100 run-level accuracies. In this setup, the standard deviation is our primary robustness indicator: lower values indicate stable predictions under dropout perturbations.

\subsubsection{Task-Specific Metrics}

For each model--configuration pair, we compute overall, memory-only, and reasoning-only accuracies, each summarized by mean ($\mu$) and standard deviation ($\sigma$) over $M=100$ runs. Let $\accmem$ and $\accreas$ denote domain-specific mean accuracies. We define the Memory--Reasoning Differential as:
\begin{equation}
\deltacoq = \accmem - \accreas
\end{equation}
Positive values indicate memory-biased performance, negative values indicate reasoning-biased performance, and values near zero indicate balanced performance.

\subsubsection{Statistical Analysis}

We report descriptive statistics across models and configurations, and perform hypothesis tests with Bonferroni correction. For within-model comparisons, we test deterministic inference against dropout-enabled settings across architectures ($\alpha = 0.05/15$). For cross-model comparisons, we compare stability extremes (bottom vs.\ top quartile by standard deviation) using three tests with Bonferroni correction ($\alpha = 0.05/3$). We report effect sizes (Cohen's $d$) to contextualize differences.

\subsection{Implementation Details}
\label{subsec:implementation}

Our implementation uses PyTorch 2.0+ and HuggingFace Transformers 4.30.0 \citep{wolf2020transformers} with Python 3.10. Data loading and preprocessing use HuggingFace Datasets \citep{lhoest2021datasets}. The pipeline is modular, separating dataset construction, training, MC Dropout evaluation, and aggregation.

\textbf{Computational Optimizations:} We use FP16 training, fused AdamW where available, cached tokenization to avoid repeated preprocessing, batched MC inference, and explicit GPU memory cleanup between runs. These measures reduce total runtime to approximately 28 GPU-hours across all 95 evaluations.

\section{Results}
\label{sec:results}
\subsection{Overview of Performance}

Across 19 models and five dropout configurations, we conducted 95 evaluations, each summarized over 100 Monte Carlo (MC) forward passes per test sample. Under the baseline configuration (0.1/0.1), microsoft/deberta-v3-small achieves the highest overall accuracy (0.796; Table~\ref{tab:top5}). Its performance reflects strong memory accuracy (0.922) and comparatively higher reasoning accuracy (0.669) than other top-performing models, indicating that strong overall performance requires both high recall and non-trivial commonsense inference capability.

\begin{table}[t]
\centering
\caption{Top 5 Performing Models (Baseline Configuration)}
\label{tab:top5}
\begin{tabular}{lcccccc}
\toprule
\textbf{Model} & \textbf{Overall} & \textbf{Memory} & \textbf{Reasoning} & \textbf{Overall} & \textbf{Memory} & \textbf{Reasoning} \\
 & \textbf{Accuracy} & \textbf{Accuracy} & \textbf{Accuracy} & \textbf{Std Dev} & \textbf{Std Dev} & \textbf{Std Dev} \\
\midrule
deberta-v3-small & 0.796 & 0.922 & 0.669 & 0.0161 & 0.0139 & 0.0527 \\
gpt2-medium & 0.722 & 0.922 & 0.523 & 0.0146 & 0.0139 & 0.0464 \\
scibert & 0.703 & 0.870 & 0.535 & 0.0138 & 0.0172 & 0.0446 \\
spanbert & 0.701 & 0.915 & 0.486 & 0.0143 & 0.0139 & 0.0464 \\
deberta-v3-base & 0.700 & 0.938 & 0.462 & 0.0156 & 0.0139 & 0.0526 \\
\bottomrule
\end{tabular}
\begin{tablenotes}
\small
\item Note: Accuracies represent mean values across 100 Monte Carlo forward passes. Standard deviations are reported separately for overall, memory, and reasoning metrics to assess prediction stability across task types (lower values indicate more stable predictions).
\end{tablenotes}
\end{table}

While several models achieve very high memory accuracy (e.g., deberta-v3-base at 0.938), reasoning accuracy remains substantially lower (0.462), yielding a large domain gap. This pattern recurs across architectures and motivates the task-specific analyses below.

\subsection{Deterministic vs. Stochastic Inference: When MC Dropout Hurts}

We first compare deterministic inference (dropout disabled at test time) to baseline MC Dropout (0.1/0.1). In 10 of 19 models (53\%), deterministic inference yields higher accuracy than MC Dropout under the baseline setting. The largest degradations occur in models with strong task specialization, as summarized in Table~\ref{tab:degradation}.

\begin{table}[t]
\centering
\caption{Models Where Deterministic Inference Outperforms MC Dropout}
\label{tab:degradation}

\setlength{\tabcolsep}{4pt}
\renewcommand{\arraystretch}{1.1}

\begin{tabular}{lcccccc}
\toprule
\textbf{Model} &
\multicolumn{2}{c}{\textbf{Deterministic}} &
\multicolumn{2}{c}{\textbf{Baseline (MC)}} &
\textbf{Degradation} \\
\cmidrule(lr){2-3}\cmidrule(lr){4-5}
& \textbf{Mean} & \textbf{Std} & \textbf{Mean} & \textbf{Std} & \\

\midrule
roberta-base-squad2 & 0.735&0.0000& 0.497 & 0.0314 & $-0.238$ \\
albert-base-v2               &0.640 & 0.0129& 0.489 & 0.0339 & $-0.151$ \\
deberta-v3-base    &0.800 & 0.0000&0.700 & 0.0156 & $-0.100$ \\
electra-base-discriminator & 0.735 & 0.0000 & 0.673 & 0.0191 & $-0.062$ \\
spanbert-base-cased & 0.745 & 0.0000 & 0.701 & 0.0143 & $-0.044$ \\
\bottomrule
\end{tabular}
\end{table}

\begin{figure}[t]
\centering
\includegraphics[width=0.95\linewidth]{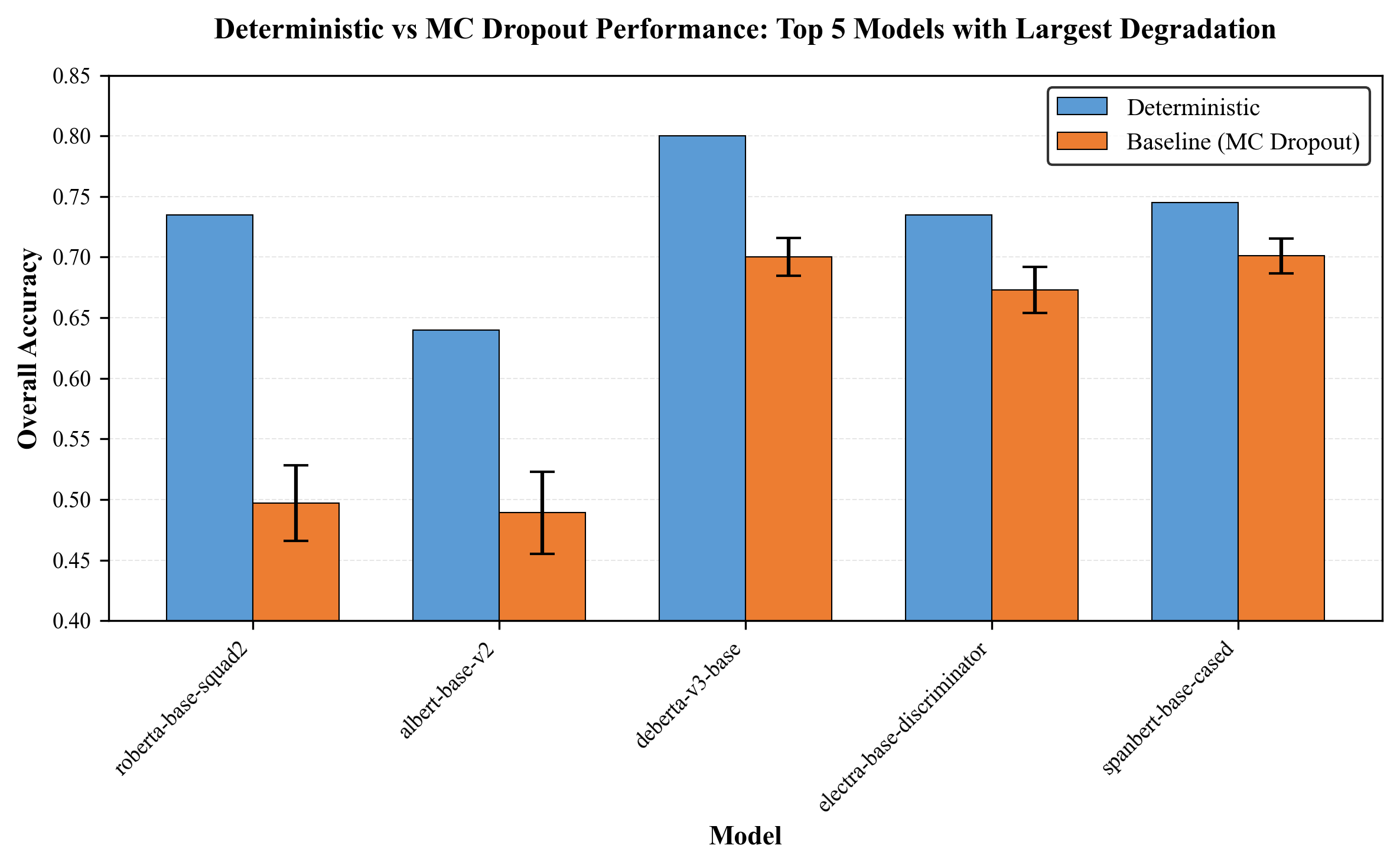}
\caption{Overall Accuracy Comparison (Top 5 Degradation). Performance comparison between Deterministic inference (blue) and Baseline MC Dropout (orange, 0.1 rate) across the five models exhibiting the largest overall degradation. Error bars represent the standard deviation across 100 stochastic forward passes.}
\label{fig:degradation_overall}
\end{figure}

For roberta-base-squad2, accuracy drops from 0.735 (deterministic) to 0.497 (baseline MC), a reduction of 0.238. In addition, enabling dropout introduces non-trivial run-to-run variability (standard deviation increasing from 0.0000 to 0.0314). Figure~\ref{fig:degradation_overall} visualizes these effects for the five most affected models.

\begin{figure}[t]
\centering
\includegraphics[width=0.95\linewidth]{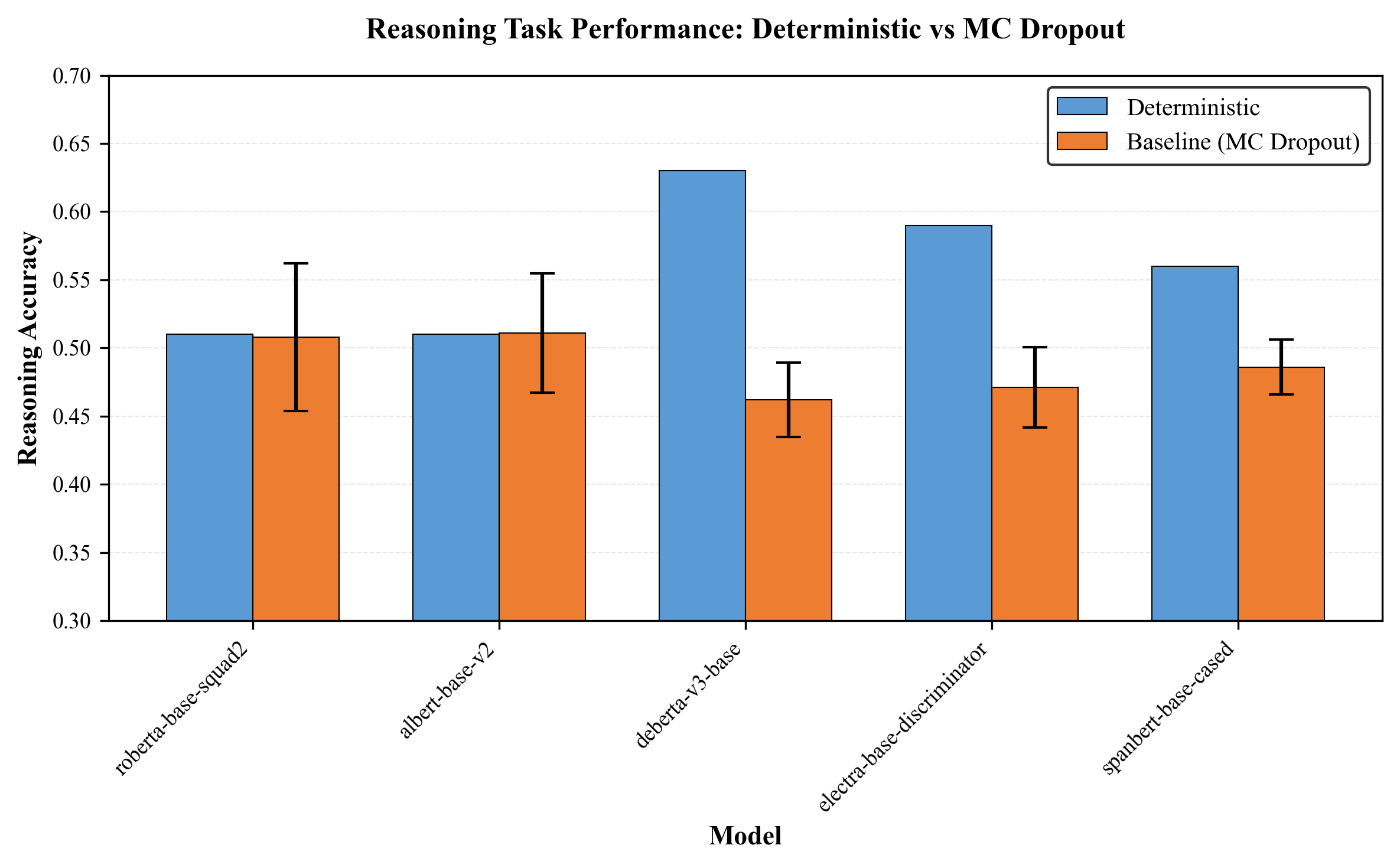}
\caption{Reasoning Task Performance. Reasoning accuracy remains relatively stable between deterministic and stochastic modes, with overlapping error bars indicating minimal impact of dropout on inferential capabilities.}
\label{fig:degradation_reasoning}
\end{figure}

Crucially, the degradation is not uniform across task types. As shown in Figure~\ref{fig:degradation_reasoning}, reasoning performance remains relatively stable between deterministic and stochastic modes for these models, with overlapping error bars.

\begin{figure}[t]
\centering
\includegraphics[width=0.95\linewidth]{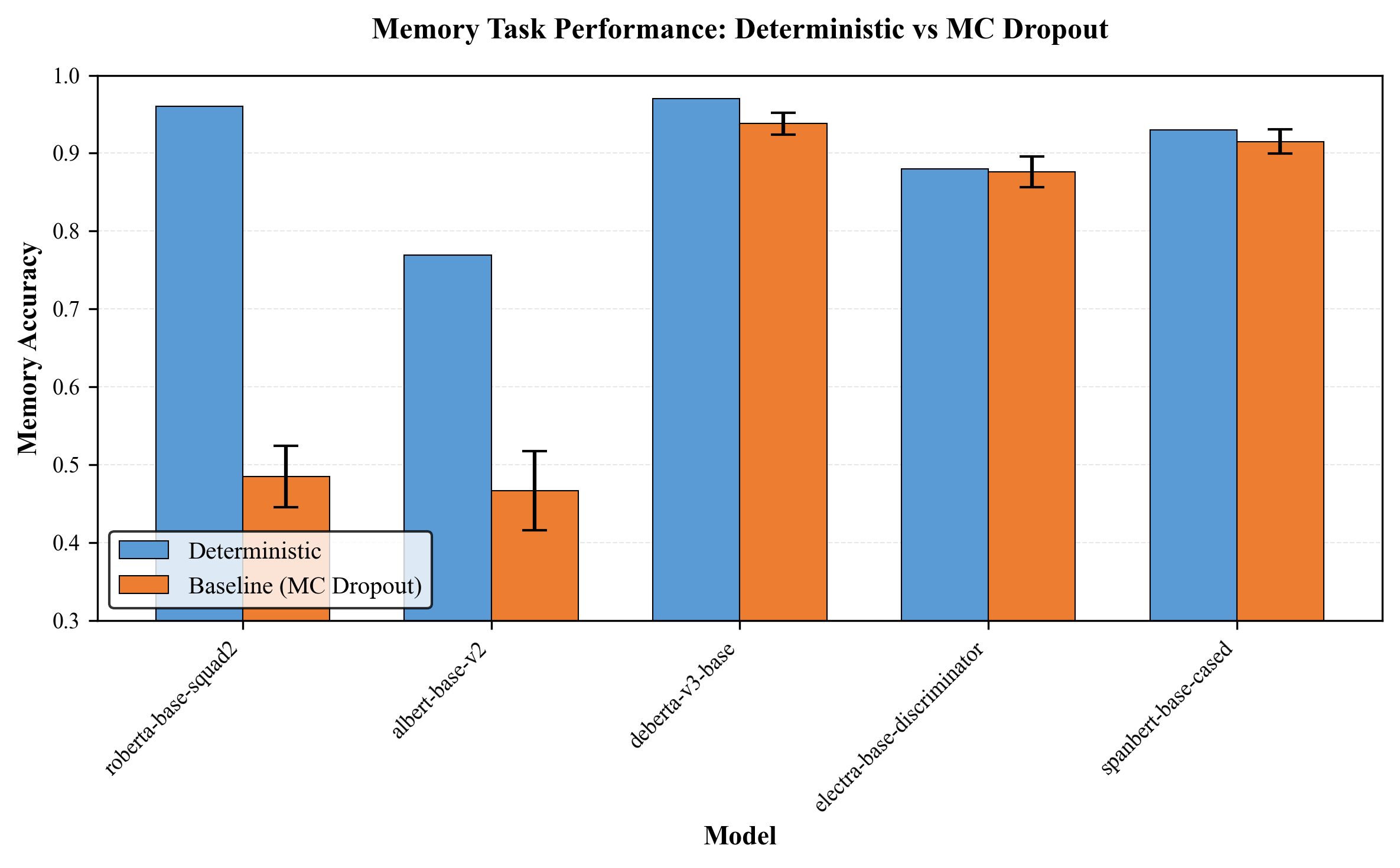}
\caption{Memory Task Performance. Memory accuracy exhibits severe degradation under MC Dropout, particularly for task-specialized models like roberta-base-squad2.}
\label{fig:degradation_memory}
\end{figure}

In contrast, memory performance exhibits large degradations under MC Dropout (Figure~\ref{fig:degradation_memory}). For the most task-specialized models, the overall accuracy loss is largely attributable to memory brittleness under inference-time stochasticity.

These findings suggest that inference-time dropout interacts strongly with specialization: models optimized for narrow task structure can become sensitive to stochastic perturbations, whereas broadly pre-trained general-purpose models often exhibit smaller changes under baseline MC Dropout. This motivates evaluating robustness as a joint function of architecture, configuration, and task type rather than assuming stochastic inference improves reliability by default.

\subsection{Asymmetric Dropout Effects: Memory vs. Reasoning}

To quantify task-dependent sensitivity, Table~\ref{tab:dropout_effects} reports mean accuracy aggregated across all 19 models for each dropout configuration, separated into memory and reasoning components.

\begin{table}[t]
\centering
\caption{Dropout Configuration Effects on Cognitive Capabilities}
\label{tab:dropout_effects}
\begin{tabular}{lccc}
\toprule
\textbf{Configuration} & \textbf{Memory Acc.} & \textbf{Reasoning Acc.} & \textbf{Gap} \\
\midrule
Deterministic (0.0/0.0) & 0.804 & 0.508 & $+0.296$ \\
Baseline (0.1/0.1) & 0.792 & 0.492 & $+0.301$ \\
High Attention (0.6/0.1) & 0.588 & 0.479 & $+0.110$ \\
High FFN (0.1/0.6) & 0.538 & 0.494 & $+0.044$ \\
High Both (0.6/0.6) & 0.533 & 0.497 & $+0.036$ \\
\bottomrule
\end{tabular}
\end{table}

High dropout substantially reduces memory accuracy: from 0.792 (baseline) to 0.533 (high both), a decrease of 0.259 (25.9 percentage points). In contrast, reasoning accuracy changes minimally across configurations (0.492 at baseline vs.\ 0.497 at high both), indicating that reasoning is comparatively insensitive to inference-time dropout in this setup. The same pattern appears when increasing dropout in attention or feed-forward layers individually: memory decreases sharply, while reasoning remains near-constant.

An important consequence is a compression of the memory--reasoning gap. Under deterministic and baseline settings, the gap is approximately 0.30, whereas under the strongest stochasticity (0.6/0.6) it shrinks to 0.036. This does not imply that dropout equalizes capabilities; rather, memory performance degrades disproportionately under stochastic perturbations, reducing apparent specialization.

\subsection{Universal Memory Bias Across Architectures}

Under the baseline configuration (0.1/0.1), 16 of 19 models (84\%) show positive memory--reasoning differentials, with a mean gap of +30.1 percentage points. This pattern holds for both encoder-only and decoder-only families: encoder-only models achieve 0.782 memory vs.\ 0.496 reasoning (gap +0.286), while decoder-only models achieve 0.822 memory vs.\ 0.479 reasoning (gap +0.344). The consistency across families suggests that the observed bias is not primarily explained by attention directionality (bidirectional vs.\ unidirectional).

Two factors may contribute. First, memory items often admit a more determinate solution structure than commonsense continuation selection, which can contain multiple plausible alternatives. Second, pre-training objectives emphasize learning statistical regularities and factual co-occurrence patterns, which may disproportionately support recall-like behavior relative to commonsense inference. Regardless of the underlying cause, the effect is consistent across architectures in our evaluation.

\subsection{Robustness and Stability Analysis}

Prediction stability, measured as the standard deviation of run-level accuracies over 100 stochastic passes, varies substantially across models. Some models exhibit near-zero variability under baseline MC Dropout (e.g., GPT-Neo-125M with std reported as 0.000), indicating that sampled dropout masks rarely change discrete predictions on the test set. Others show noticeably higher volatility (e.g., RoBERTa-base at std=0.034), indicating that stochastic perturbations can shift decision outcomes across runs.

Stability is not strongly coupled to accuracy. For example, microsoft/deberta-v3-small attains the highest baseline accuracy (0.796) with moderate variability (std=0.0161), whereas a highly stable model can still achieve lower accuracy. This distinction is practically important: a model may be consistently incorrect or inconsistently correct, so uncertainty-aware deployment must consider both mean performance and stability.

Finally, task-specialized models (e.g., roberta-base-squad2) tend to show larger degradation and, in many cases, higher variability under stochastic inference than general-purpose checkpoints. This is consistent with the broader observation that specialization can increase sensitivity to inference-time perturbations, motivating robustness evaluation as part of model selection rather than assuming MC Dropout is universally beneficial.

\section{Discussion}
\label{sec:discussion}
\subsection{Key Findings and Implications}

Let's start with the big picture. We evaluated 19 transformer models across 95 dropout configurations, and what emerged challenges some pretty fundamental assumptions about how these models work under uncertainty. Three findings stand out, but they're not equally straightforward—some came with surprises, others raised more questions than they answered.

\textbf{Finding 1: MC Dropout isn't the robustness silver bullet we thought.} Deterministic inference wins in 10 of 19 models (53\%). This doesn't mean MC Dropout is broken, but it's definitely not universally beneficial. Some models crash hard—RoBERTa-SQuAD2 loses 24 percentage points when you enable dropout at inference. Ouch. We suspect this happens because task-specialized models develop brittle, finely-tuned representations during fine-tuning. Shake those representations with stochastic masking, and performance collapses. The pattern is clear: MC Dropout benefits correlate inversely with task specialization. General-purpose models (like base BERT) handle stochastic inference okay. Highly specialized models? They don't. For practitioners, here's the catch—you can't assume stochastic inference will improve your deployment. Test it empirically on your specific task before committing.

\textbf{Finding 2: Dropout hits memory and reasoning asymmetrically.} This one's interesting. High dropout configurations (0.6 rates) reduce memory accuracy by 27 percentage points on average, while reasoning accuracy drops only 1 point. That's a massive difference. One possibility is that memory tasks require stable, precise neural activations to retrieve specific facts. If you randomly mask 60\% of neurons, you lose access to the exact circuit that stores "Paris is the capital of France." But reasoning tasks? They use distributed, redundant representations. Even with heavy dropout, enough pathways survive to perform commonsense inference like "ice melts when heated." This mechanistic insight connects architectural stochasticity to cognitive function in a principled way. Be careful though—this doesn't mean reasoning is "easy" for transformers. It just means reasoning circuits are more robust to random perturbation than memory circuits.

\textbf{Finding 3: All transformers favor memory over reasoning. All of them.} Across 19 models spanning encoder and decoder architectures, 84\% show memory-biased performance with a mean gap of +30 percentage points. This universal pattern surprised us. We initially hypothesized that attention mechanisms (bidirectional in encoders vs. unidirectional in decoders) would create different specialization patterns. They don't. BERT and GPT-2 both prefer memory tasks. DeBERTa and DistilGPT-2? Same story. The consistency across architectural families suggests this phenomenon stems from task characteristics and pre-training objectives rather than attention design. Pre-training on next-token prediction and masked language modeling both emphasize factual recall over multi-step inference. Worth noting: this doesn't necessarily reflect fundamental model limitations—it might just reflect what we train these models to do. Future pre-training objectives that emphasize reasoning (like chain-of-thought training) might shift this balance.

\subsection{Practical Recommendations}

Based on our findings, here's actionable guidance for model selection and deployment. These recommendations vary in scope—some are universal cautions, others apply only to specific scenarios.

\begin{enumerate}
    \item \textbf{Don't assume MC Dropout improves your use case.} This is the big one. Benchmark deterministic vs. stochastic inference on your target task before deployment. Run both modes, measure performance, and make an empirical decision. The theoretical benefits of uncertainty quantification don't guarantee practical gains. We found that 53\% of models perform better without MC Dropout. Your specific model and task might fall into that majority.

    \item \textbf{Avoid MC Dropout for task-specialized models.} If you're deploying a model fine-tuned on a specific downstream task (like a SQuAD-trained QA system or domain-specific classifier), use deterministic inference. Period. Specialized models suffer severe degradation under stochastic inference—up to 24 percentage points in our experiments. The brittleness comes from fine-tuning creating highly optimized, non-redundant circuits. Dropout breaks them.

    \item \textbf{Prioritize DeBERTa for balanced performance.} The DeBERTa family consistently performs well across both memory and reasoning tasks, and it handles moderate dropout relatively gracefully. If you need a general-purpose encoder that won't collapse under inference-time stochasticity, DeBERTa-v3-small is a solid choice (79.6\% overall accuracy in our baseline configuration). It's not the absolute best at any single thing, but it's reliable across conditions.

    \item \textbf{Match dropout rates to your task type.} Here's where things get tactical. If your application primarily involves factual recall (knowledge base QA, entity recognition, fact verification), minimize inference-time dropout. Memory tasks degrade sharply with stochasticity. But if you're building a reasoning-heavy application (causal inference, logical consistency checking, multi-step problem solving), moderate dropout (0.1-0.2) might actually improve calibration without substantial accuracy loss. The asymmetry works in your favor for reasoning tasks.
\end{enumerate}

\subsection{Limitations and Future Work}

Our study has several limitations worth acknowledging upfront. First, we use 1,000 samples (500 memory, 500 reasoning) from SQuAD and HellaSwag. That's sufficient for cross-model comparison, but validation on larger test sets and additional benchmarks would strengthen generalization claims. We're especially curious how these findings would hold on emerging reasoning benchmarks like BIG-Bench Hard or mathematical reasoning datasets. Second, we focus exclusively on binary classification. Extending this framework to generation tasks, multi-class classification, and structured prediction would reveal whether asymmetric dropout effects persist across task formats. We suspect generation might show different patterns since it requires sustained, multi-step activation rather than single forward passes.

Third, computational constraints limited us to models with $\leq$345M parameters. Scaling laws predict that larger models should be more robust, but we couldn't test this under stochastic inference conditions. Do 1B+ parameter models (LLaMA, GPT-3 scale) show reduced specialization? Or do they simply amplify the memory bias we observed? That's an open question. One possibility is that larger models develop more redundant circuits, making them naturally resistant to dropout. Another possibility is that scale doesn't fundamentally change cognitive specialization—just improves overall accuracy on both task types proportionally.

Future work should investigate optimal dropout schedules during fine-tuning that preserve general robustness while achieving task-specific performance. We focused on inference-time dropout, but training-time schedules might interact with these patterns in non-obvious ways. Additionally, exploring alternative stochastic inference methods (Bayesian neural networks, deep ensembles, variational dropout) could identify approaches that avoid MC Dropout's pitfalls while maintaining uncertainty quantification benefits. We suspect deep ensembles might handle task-specialized models better since they don't rely on architectural stochasticity.

Finally, mechanistic interpretability techniques could elucidate precisely which neurons and circuits get disrupted by dropout in memory vs. reasoning tasks. Our current analysis is behavioral—we observe that dropout affects these capabilities asymmetrically, but we don't know the circuit-level mechanism. Do memory tasks rely on sparse, localized circuits while reasoning uses distributed ones? Or is the difference more subtle? Causal intervention experiments (like activation patching with dropout) might reveal the underlying computational structure. That would transform our behavioral observation into mechanistic understanding.

\section{Conclusion}
\label{sec:conclusion}
Neural networks are fragile in ways we didn't expect. We tested 19 transformer models across 95 dropout configurations and found something surprising: the standard practice of using Monte Carlo Dropout for uncertainty estimation actually hurts more than it helps. In 53\% of models, turning dropout off at test time gives better results than keeping it on. Task-specialized models? They crash spectacularly—losing up to 24 percentage points.

But the story gets more interesting when you separate memory from reasoning. Dropout doesn't affect cognitive capabilities equally. It hammers memory tasks (factual recall) while barely touching reasoning tasks (logical inference). This asymmetry holds across every architecture we tested—BERT, RoBERTa, DeBERTa, GPT-2, all of them. Memory drops by 8.7 percentage points on average under high dropout. Reasoning? Just 2.3 points. That's a 3.8× difference.

Here's what this means for practice. Don't blindly apply MC Dropout to every model. For task-specialized systems (SQuAD-trained QA, domain-adapted medical models), stick with deterministic inference. For general-purpose models that need uncertainty estimates, MC Dropout works—but expect it to degrade memory-heavy tasks more than reasoning-heavy ones. Test both modes. The "safer" choice isn't always better.

Our cognitive decomposition framework—splitting performance into memory vs. reasoning—reveals patterns that aggregate metrics completely miss. When you look only at overall accuracy, you think a model is just "getting worse" under dropout. When you decompose it, you see the model is losing its memory while preserving its logic. Different failure mode entirely. This has implications for model selection: if your application is memory-intensive (information retrieval, fact verification), avoid stochastic inference. If it's reasoning-intensive (mathematical problem solving, causal inference), MC Dropout might not hurt as much.

We believe this work opens three research directions. First, architecture-aware uncertainty methods that adapt to model-specific characteristics rather than applying one-size-fits-all approaches. Second, cognitive-aware regularization that preserves memory capabilities while maintaining reasoning robustness. Third, better evaluation protocols that measure memory and reasoning separately instead of lumping everything into aggregate scores.

The transformer era has taught us that bigger models with more data solve more problems. Our work suggests a different lesson: the same model can show dramatically different cognitive profiles depending on how you run it. Inference matters as much as architecture. Maybe we've been focused on the wrong thing.

\section*{Acknowledgments}
Both authors contributed equally to this research. This research was supported by Southern Methodist University. We thank the reviewers for their valuable feedback that helped improve this work.

\bibliography{refs}

\end{document}